\documentclass[10pt,twocolumn,letterpaper]{article}

\usepackage{cvpr}
\usepackage{times}
\usepackage{epsfig}
\usepackage{graphicx}
\usepackage{amsmath}
\usepackage{amssymb}
\usepackage{physics}
\usepackage{commath}

\usepackage[breaklinks=true,bookmarks=false]{hyperref}

\cvprfinalcopy %

\def\cvprPaperID{5495} %

\begin{document}

\title{Multi-Path Region Mining For Weakly Supervised 3D Semantic Segmentation on Point Clouds}

\author{Jiacheng Wei\textsuperscript{1} \qquad Guosheng Lin\textsuperscript{1}\thanks{Corresponding author: G. Lin (e-mail: {\tt gslin@ntu.edu.sg})} \qquad Kim-Hui Yap\textsuperscript{1} \qquad Tzu-Yi Hung\textsuperscript{2} \qquad Lihua Xie\textsuperscript{1} \\
\textsuperscript{1}Nanyang Technological University, Singapore \quad \textsuperscript{2}Delta Research Center, Singapore \\
\small \{\tt jiacheng002, gslin, ekhyap, elhxie\}@ntu.edu.sg, \tt tzuyi.hung@deltaww.com }

\maketitle
\vspace{-20pt}
\begin{abstract}
   Point clouds provide intrinsic geometric information and surface context for scene understanding. Existing methods for point cloud segmentation require a large amount of fully labeled data. Using advanced depth sensors, collection of large scale 3D dataset is no longer a cumbersome process. However, manually producing point-level label on the large scale dataset is time and labor-intensive. In this paper, we propose a weakly supervised approach to predict point-level results using weak labels on 3D point clouds. We introduce our multi-path region mining module to generate pseudo point-level label from a classification network trained with weak labels. It mines the localization cues for each class from various aspects of the network feature using different attention modules. Then, we use the point-level pseudo labels to train a point cloud segmentation network in a fully supervised manner. To the best of our knowledge, this is the first method that uses cloud-level weak labels on raw 3D space to train a point cloud semantic segmentation network. In our setting, the 3D weak labels only indicate the classes that appeared in our input sample. We discuss both scene- and subcloud-level weakly labels on raw 3D point cloud data and perform in-depth experiments on them. On ScanNet\cite{dai2017scannet} dataset,  our result trained with subcloud-level labels is compatible with some fully supervised methods. 
   \vspace{-15pt}
\end{abstract}
\section{Introduction}
Compared to 2D images, as projections of the real world, 3D data brings geometry and the surrounding context of objects and scenes along with their RGB information. The extra hints have drawn much attention recently. With the great success of deep learning in 2D image vision tasks, researchers proposed many deep learning based methods for recognition tasks on point clouds. However, deep learning based methods are usually data-hungry.

Recently, the advances in reconstruction algorithms and more affordable consumer-level depth sensors provide convenient and inexpensive access to 3D data collection. However, annotations on these data are still expensive in labor and time. Especially for 3D data, directly labeling of reconstructed 3D meshes or grouped points is to be carried out.

\begin{figure}[t]
\begin{center}
   \includegraphics[width=0.8\linewidth]{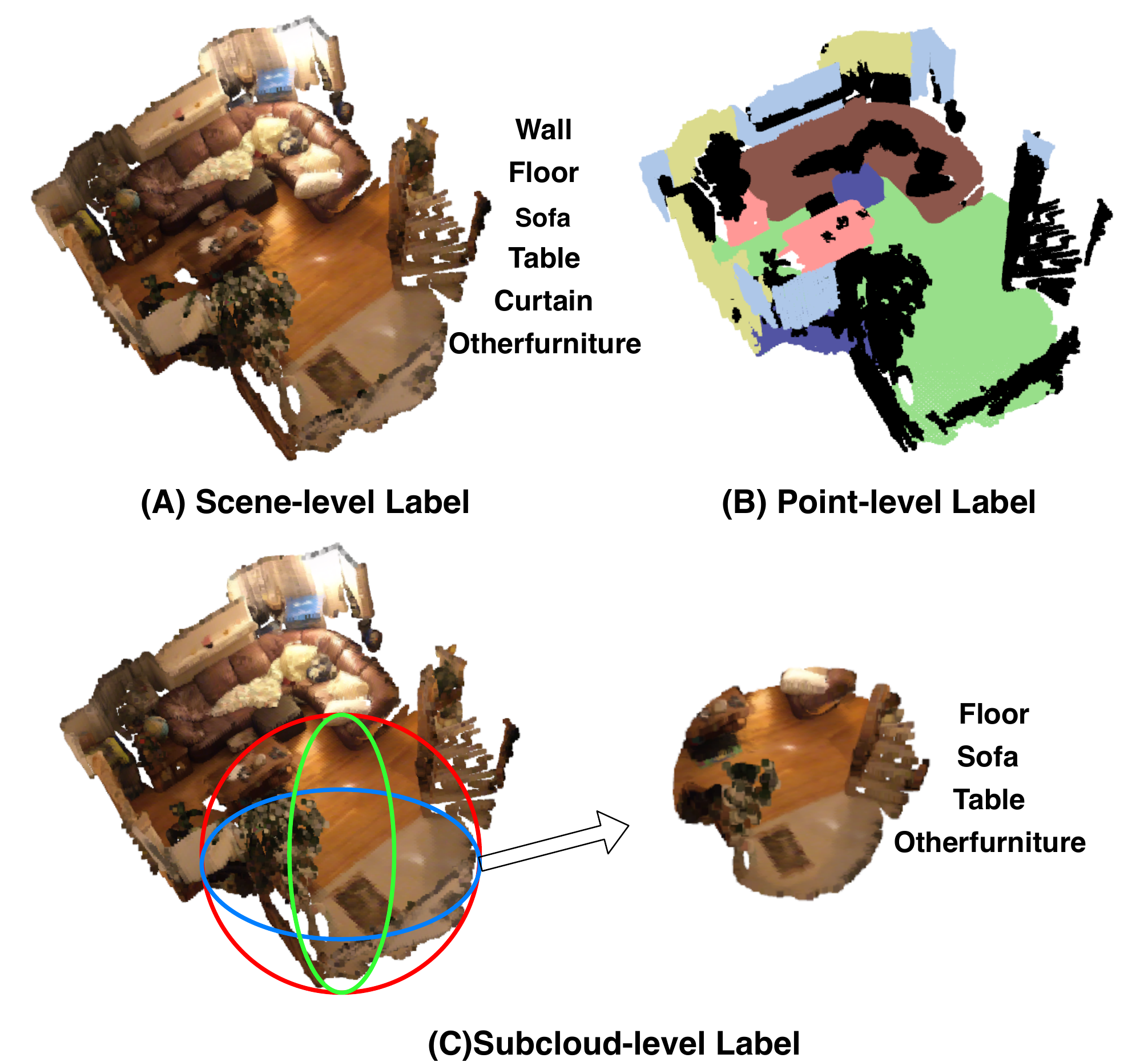}
\end{center}
   \caption{An illustration of different kinds of labels for point clouds. (A) Scene-level label indicates all the classes appeared in the scene, (B)Point-level label indicates the class that each pixel belongs to, (C)Subcloud-level label denotes the classes appeared in the subcloud.}
\label{fig:image1}
\vspace{-10pt}
\end{figure}

For example, ScanNet\cite{dai2017scannet}, a popular large-scale RGB-D dataset of real-world indoor environments, that provides 1513 3D scans from 70 unique indoor scenes, consists of over 2.5M RGB-D images. Then, 3D meshes and point clouds were reconstructed from the RGB-D scans. Using Structure sensor\cite{sensor2016occipital}, which can be attached to mobile devices like smartphones and iPads, only 20 people participated in the collection of 1513 3D scans. Despite the convenience in data collection, the annotation process turned out to be much exhausting and arduous. According to the statistics, more than 500 works participated in the semantic annotation process. To ensure the annotation accuracy, each scene was annotated by 2 to 3 participants. In aggregate, the median and mean time for annotation per scan is 16.8 min and 22.3 min.

In this paper, we introduce a weakly supervised learning approach for 3D point cloud semantic segmentation. To reduce the labor cost in data annotation, we use weak labels that only indicate the classes that appeared in the input point cloud sample. Therefore, we can only train a classification network with weak labels. To find object regions for the classification network, we introduce Class Activation Map (CAM)\cite{zhou2016learning}, an effective class-specific region localization method in 2D images, to 3D point clouds. However, CAM often works well only in most discriminative regions. As we want to generate accurate pseudo labels on all points in the point cloud, we propose a multi-path region mining (MPRM) module to mine various region information from a classification network trained with weak labels. In our MPRM module, we append various attention modules after the classification backbone network. We propose a spatial attention module to gather long-range context information along the spatial domain, a channel attention module to explore channel inter-dependencies, and a point-wise spatial attention module to aggregate global information to local features. Each attention module produces a classification prediction and is trained with the weak label. By applying Point Class Activation Map (PCAM) to each path and the original feature map, we can collect object regions mined from various aspects of the network feature and generate our point-level pseudo mask. To leverage the low-level representation and the pairwise relationship of point clouds, we use a denseCRF\cite{krahenbuhl2011efficient} to refine our pseudo labels. Finally, we train a point cloud segmentation network using our point-level pseudo labels.

While weak labels are cheap to acquire, they may be too poor to provide supervision for the network to generate localization cues. To find a trade-off between labor cost and representation ability, we discuss two weak labeling strategies. In Figure \ref{fig:image1}, we show (A) scene-level label, indicating the classes appearing in the scene, which is the cheapest label we can get for a point cloud scene; (B) the commonly used point-level label; and (C) subcloud-level label, where we take spherical subsamples from the scene and label it with classes appearing in the subcloud. To ensure the labor cost remains low for subcloud-level labels, we choose only a limited number of subclouds for each scene. In ScanNet, the average number of subclouds is 18.4. The estimated annotation time for scene-level labels in a scene is around 15 sec, while the annotation time for subclouds from a scene is lower than 3 min, which is still much cheaper than point-level annotation.

We perform detailed experiments of our MPRM using both scene-level labels and subcloud-level labels. We elaborate that our approach provides a feasible way to train a point cloud segmentation network using weak labels. Our result outperforms some popular fully supervised point cloud recognition models like PointNet++\cite{qi2017pointnet++} and \cite{su2018splatnet}. Also, we show that the model trained using subcloud-level labels outperforms the model trained with scene-level labels by a large margin.

The main contribution of this paper can be summarized as:
\begin{itemize}
\item We propose a weakly supervised learning approach for 3D point cloud semantic segmentation tasks using only scene- and subcloud-level labels. To the best of our knowledge, this is the first approach to learn a point cloud scene segmentation network from cloud-level weak labels on raw 3D data.
\item We propose a multi-path region mining module to generate pseudo point-level labels. With the spatial attention module, channel attention module and point-wise spatial attention module in our MPRM, we mine various object localization region cues by exploring the long-range spatial context, channel inter-dependency, and global context from the network feature.
\end{itemize}

\section{Related Work}

\textbf{Weakly Supervised Semantic Segmentation on 2D images:} Various kinds of supervision have been studied to relieve the labor cost for dense annotation on images. Weak annotations like Bounding box\cite{khoreva2017simple,song2019box}, scribble\cite{lin2016scribblesup}, point\cite{bearman2016s} are adopted in segmentation tasks. While these kinds of supervision still require a certain amount of labor cost, image-level annotation is much cheaper. A common practice for image-level supervision tasks is generating Class Activate Map (CAM)\cite{zhou2016learning}. The core idea is extracting localization cues from a classification network for each class. Then, a segmentation network is trained with the CAM as pseudo labels. However, as CAM often fails to find the entire object region, many works\cite{kolesnikov2016seed,huang2018weakly,krahenbuhl2011efficient,wei2017object,ahn2018learning,fan2018associating}  are proposed to improve the accuracy of pseudo labels. Although there is quite a number of weakly supervise approaches for image segmentation, it is hard to directly apply them to point cloud due to its unordered structure and variant density.

\textbf{Deep Learning on Point Clouds:} To apply deep learning techniques on point clouds, some methods project 3D point clouds to images and process them on 2D \cite{boulch2017unstructured, su2015multi,tatarchenko2018tangent}, but this kind of methods often suffers a lot of deficiencies in segmentation tasks due to occlusion and variant densities. It is also popular to voxelize point clouds into 3D grids and process them using dense 3D CNNs\cite{maturana2015voxnet, ben20183dmfv}. As 3D CNNs consume huge computational resources, sparse convolutions use hash-maps\cite{graham20183d,choy20194d} to improve performance and efficiency for voxel-based methods. To reduce the quantization effort, PointNet like methods\cite{qi2017pointnet,qi2017pointnet++,liu2019point2sequence,li2018so} are proposed to directly process the raw unordered point cloud data. This kind of approach is weak in considering neighboring local information. Point convolution networks\cite{atzmon2018point,hua2018pointwise,groh2018flex,xu2018spidercnn,thomas2019kpconv,mao2019interpolated,cai2019deep,hou20193d,hu2018semantic} introduce convolution operations directly onto raw point cloud data. However, the above methods are all trained with full supervision, thus require a lot of fully annotated data.

\textbf{Point Cloud Recognition with Less Supervision:} \cite{sauder2019selfsupervised} proposes a self-supervised method to learn a point cloud representation by reassembling randomly split point clouds parts. MortonNet\cite{thabet2019mortonnet} uses Z-order to learn a feature with self-supervision. However, these two models cannot directly use the self-supervised learned feature for tasks like object classification, part segmentation, and semantic segmentation. Pre-training the network with their learned features can help improve the performance and use less fully annotated labels, which turns the problem into a semi-supervised setting. \cite{wang2019weakly3d} proposes to use 2D semantic annotation for 3D point cloud semantic segmentation tasks by re-projecting segmentation predictions on the 3D point cloud to 2D. However, it requires dense 2D annotation, which is still expensive. Thus, these methods still require quite an amount of expensive annotation, and there is no existing method that directly uses weak 3D labels for 3D scene segmentation tasks.

\section{Our Weakly Supervised Setting}\label{sec:weaklabel}
In this section, we introduce and discuss scene-level weak labels and subcloud-weak labels for our weakly supervised setting. 

\textbf{Scene-level Annotation:} Among weak labels for 2D images, image-level labels are the cheapest one to produce. In the 3D case, scene-level labels are also the most economical ones. It only indicates the appearing classes in each scene. Although researchers developed many successful approaches on 2D weakly supervised segmentation using image-level labels, there are two major challenges for using scene-level labels in 3D weakly supervised scene segmentation:
(1)3D data are reconstructed from RGB-D sequences, which usually contain much more information than a single image. Thus, a single label for a large scene is considerably coarse; (2)For indoor scenes, there are several common classes appear in high frequencies. Classes like walls and floors appear in almost every scene, and they usually have a dominant number of points in each scene. With this severe class imbalance issue, the classification network may not be able to learn the discriminative features, which makes it hard for us to find class region localization cues.

\textbf{Subcloud-level Annotation:} To deal with the above challenges while retaining the low annotation cost, we propose subcloud-level labels for indoor scene point cloud data. %
We uniformly place seeding points in the space and take the all the neighboring points within a radius $r$ to form a subcloud. The number of seeding points along each dimension is calculated as $\mathnormal{n}= \left \lceil{\mathnormal{l} / \mathnormal{r}}\right \rceil $. So the total number of seeding points in the 3D case is $m_i=n_x\times n_y\times n_z$. The subclouds are overlapping with each other. Thus each point can be included in multiple subclouds. Given a seeding point $\mathbf{q}$, a subcloud can be expressed as:
\begin{equation}\label{eq:subcloud}
    \mathcal{N}(\mathbf{q},\mathnormal{r}) = \{\mathbf{p} \in \mathcal{P} \mid  \norm{\mathbf{p}-\mathbf{q}}  < \mathnormal{r} \},
\end{equation}
where $r$ is the radius of the subcloud and $\mathbf{p}$ is taken from the set of points $\mathcal{P}$ of the entire point cloud scene. Then, human annotators can label it as the classes appeared in the subcloud. In ScanNet, using a radius $r=2m$, the average number of subclouds in each scene is 18.4. So the annotators only need to append about 18 label vectors to each scene. We present a detailed discussion on how subcloud-level labels can help resolve the challenges of directly using scene-level in our weakly supervised training scheme in \ref{subexp}.

\begin{figure}[b]
\vspace{-10pt}
\begin{center}
   \includegraphics[width=0.8\linewidth]{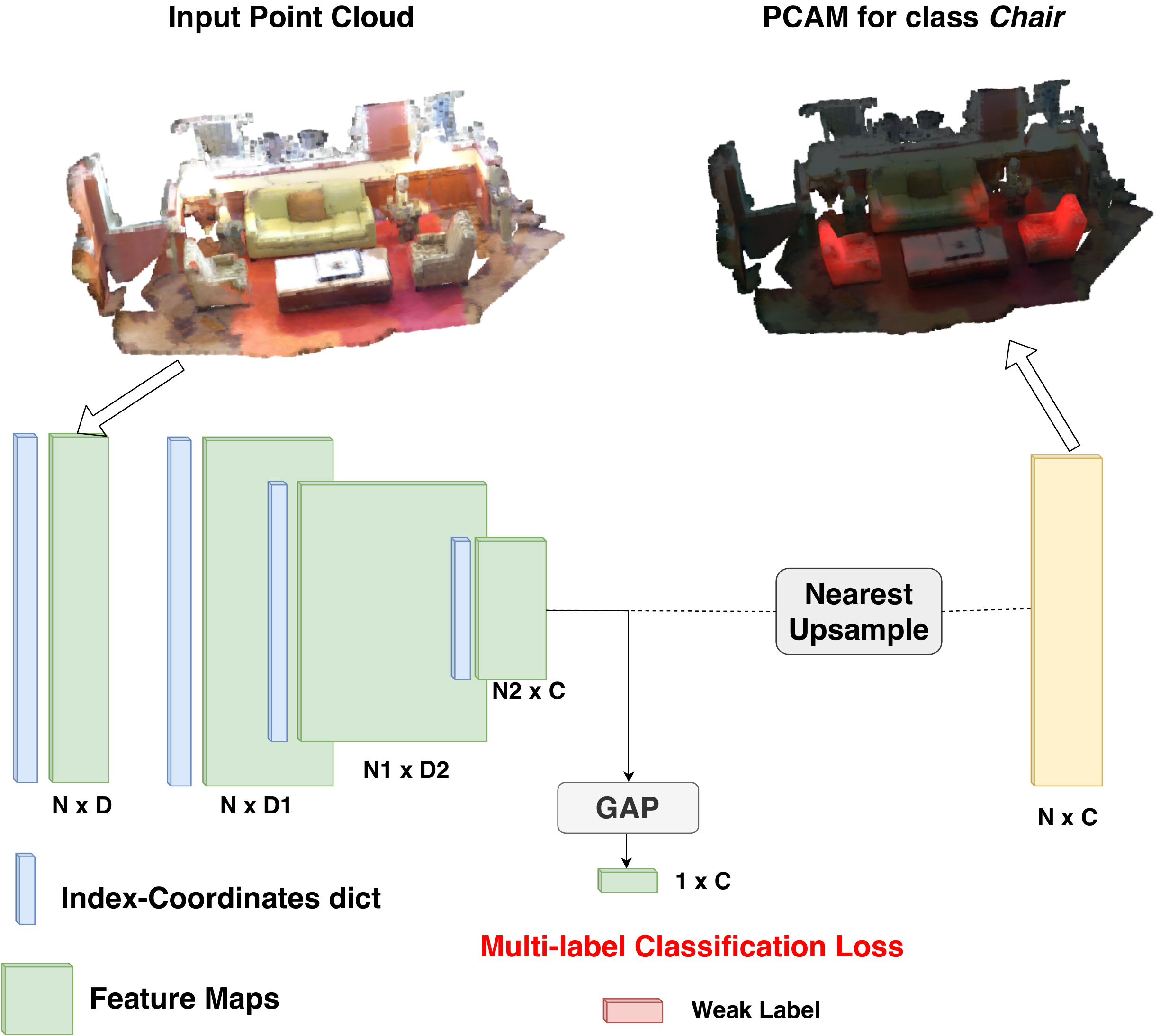}
\end{center}
   \caption{An illustration of point class activation map, we show the network architecture to generate PCAM using weak labels and visualization of PCAM for class \textit{Chair} of the given input point cloud. We can see the highly activated points in the PCAM lead us to locate the chairs region. (Best viewed in color)}
\label{fig:pcam}
\end{figure}

\begin{figure*}[t]
\begin{center}
\includegraphics[width=0.9\linewidth]{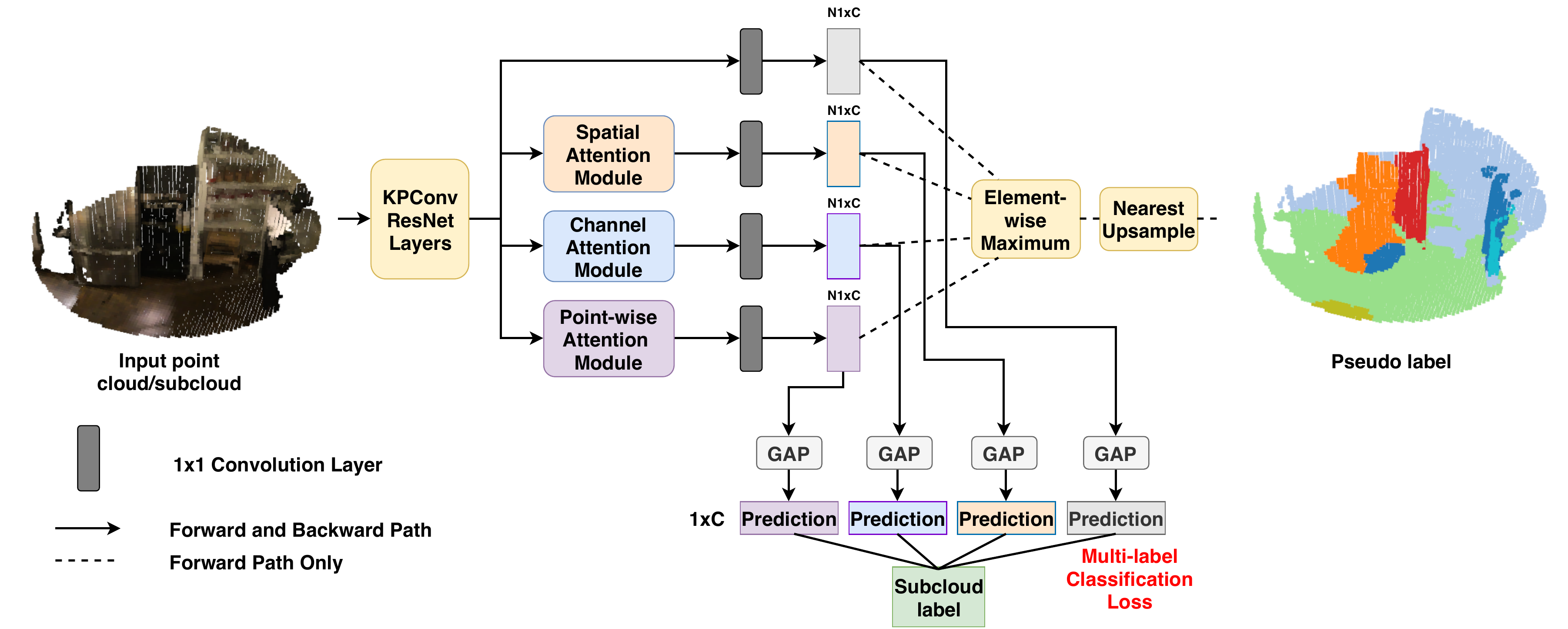}
\end{center}
   \caption{The procedure of pseudo label generation using our multi-path region mining module. We feed the input point cloud/subcloud to the network. Then we use four paths after the backbone network. Each path is a classification head with a different attention module. We take the PCAMs from each path and apply an element-wise maximum on them to get the pseudo labels.}
\label{fig:mb}
\vspace{-15pt}
\end{figure*}

\section{Our Framework}

\subsection{Baseline Method: PCAM}\label{sec:pcam}
CAM\cite{zhou2016learning} plays a vital role in weakly supervised semantic segmentation tasks on 2D images as a class-specific object localization tool. We present a point class activation map (PCAM), in which we apply CAM to point convolution networks to generate localization cues on point cloud data.

We use KPConv\cite{thomas2019kpconv} classification network with ResNet\cite{he2016deep} blocks as our backbone network. KPConv is a point convolution network that directly takes unordered points as input. It proposes a kernel convolution operation that performs convolution on a point with all its neighbors within the kernel radius in 3D space using an index-coordinate dictionary. As shown in Figure \ref{fig:pcam}, we feed point cloud/subcloud and the corresponding weak label to the classification network. Then, we take the output features map from the convolution layers. A $1 \times 1$ convolution layer is appended as a classifier to reduce the feature dimension to the number of classes to get the PCAM feature map. During training, we take the prediction vector using the global average pooling layer and calculate a sigmoid cross-entropy loss using the weak label. 

We denote $f_{cam}(p)$ as the PCAM feature vector of point $p$ before the global average pooling layer. For class $c$, the PCAM $M_c(p)$ for point $p$ can be expressed as:
\begin{equation}
    M_{c}(p) = \mathbf{w}_{c}^\intercal \cdot f_{cam}(p) \cdot  \mathbf{y}_{c},
\end{equation}
where $\mathbf{w}_{c}$ is the classification weights for class $c$ and $\mathbf{y}_{c} \in \{0,1\}$ indicates the one-hot subcloud ground truth for class $c$. Thus, we ensure that the PCAMs for negative classes in our subcloud label are set to 0. Since there is no absolute background in point cloud data, every point must be classified to a foreground class, the pseudo label of point $p$ is given by $argmax(M(p))$. Since there are several downsampling operations in the classification network, we need to upsample the PCAMs to the original scale for the point-level pseudo labels.

\subsection{Multi-Path Region Mining}
In our weakly supervised learning framework, we train a classification network with classification labels and try to find class region localization cues from the network. However, a classification network is trained only to predict class labels for the input point cloud. Learning from the most discriminative features is enough for the classification tasks. Thus, it is hard to determine class information using PCAMs at non-discriminative regions. Therefore, we want to mine more discriminative regions from the network using various kinds of attention mechanisms. Since each attention mechanism focuses on different aspects of the network features, we can produce different discriminative regions and aggregate them to generate our point-level pseudo labels.

As shown in Figure \ref{fig:mb}, our multi-path region mining module consists of four different paths after the KPConv ResNet layers. The first path is a plain PCAM introduced in \ref{sec:pcam}. In parallel, we have spatial attention module, channel attention module, and point-wise attention module. Each path is followed by a $1\times1$ convolution layer as a classifier to generate an individual PCAM. Then, we use a global average pooling layer to generate the prediction vectors and calculate a sigmoid cross-entropy loss using the weak label for each path. All the losses will be back-propagate to the backbone network. In order to generate the pseudo-label, we take the PCAMs from each path and merge them by taking the element-wise maximum value and upsample the PCAM to the original size by nearest upsampling. By taking the maximum value, we can collect discriminative features from different paths with various aspects of the classification network. Thus, we can produce more accurate point-level pseudo labels.

\begin{figure}[t]
\begin{center}
   \includegraphics[width=0.8\linewidth]{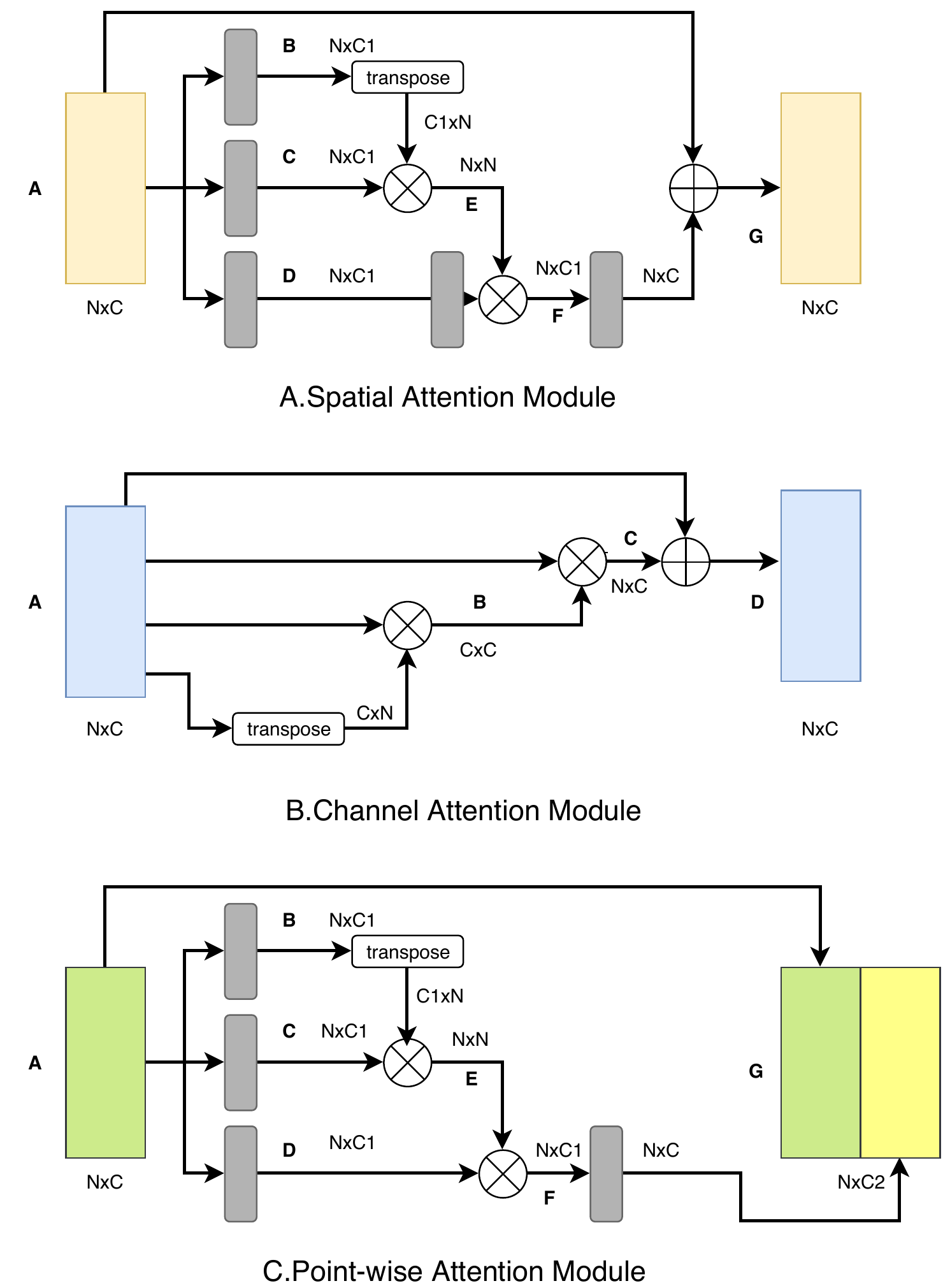}
\end{center}
   \caption{The architecture of (A) Spatial Attention Module, (B) Channel Attention Module and (C) Point-wise Attention Module}
\label{fig:attention}
\vspace{-10pt}
\end{figure}
\vspace{-10pt}
\subsubsection{Spatial Attention Module}
To determine more accurate object regions, we want to generate more discriminative features from the classification network. However, \cite{zhao2017pyramid} suggests that local features generated by traditional CNN models may mislead the classification process. Inspired by \cite{wang2018non}, we introduce a spatial attention module as a path with a classification head. By propagating the long-range context information along spatial domain, this module help ameliorate the local discriminative feature representation, which allows a more accurate object region localization procedure.

As shown in Figure \ref{fig:attention}(A), we take the $N\times C$ output feature map $\mathbf{A}$ from the backbone network and feed them to three distinct $1\times 1$ convolution layers. The dimension of the outputs $\mathbf{B}$, $\mathbf{C}$, $\mathbf{D}$ are reduced to $C1$. Then, we take the matrix multiplication between $\mathbf{C}$ and the transpose of $\mathbf{B}$ to get a $N \times N$ output matrix and perform softmax on it to get the spatial attention map $\mathbf{E}$. The attention value on $\mathbf{E}$ can be expressed as:
\begin{equation}
    e_{ji} = \frac{exp(C_i \cdot B_j)}{\sum_{i=1}^{N} exp(C_i \cdot B_j)},
\end{equation}
where $e_{ji}$ describes the influence from point $i$ to point $j$. Then, we take the matrix multiplication between the $N \times N$ spatial attention map $\mathbf{E}$ and $N \times C1$ feature map $\mathbf{D}$. Hence, we get a $N \times C1$ aggregated feature map $\mathbf{F}$. After another $1 \times 1$ convolution layer to enlarge the dimension, we use an element-wise sum to fuse the aggregated feature map $\mathbf{F}$ to the original feature map. Following \cite{zhang2018self,fu2019dual}, we use a scale parameter $\alpha$ to weight the aggregated feature $\mathbf{F}$ during summation, where $\alpha$ is a learnable parameter initialized at 0. The output of the spatial attention module as point $j$ can be expressed as:
\begin{equation}
    G_j = \alpha \sum_{i=1}^{N}(e_{ji}D_i)+A_j.
\end{equation}
It can be seen that for each point, a weighted sum of all the features from other points is added to the local feature, which selectively increased the global context to the local feature.

\subsubsection{Channel Attention Module}
In our MPRM module, we want to mine object regions from different angles of the network feature. Inspired by \cite{fu2019dual}, we introduce a channel attention module as another path with a classification head. In this module, we explore the interdependencies between channels as each channel can be represented as a class-specific response. We leverage the channel-wise information for classification, where more useful features can be mined to generate object region proposals.

As illustrated in Figure \ref{fig:attention}(B), the overall structure of the Channel Attention Module is similar to the Spatial Attention Module. However, we do not include the $1\times 1$ convolution layer in this module and directly perform matrix multiplication between the $C\times N$ transposed input feature map and the $N \times C$ original feature map. Here we get a $C\times C$ channel attention map. With a softmax layer on the attention map, it can be expressed as:
\begin{equation}
    b_{ji} = \frac{exp(A_i \cdot A_j)}{\sum_{i=1}^{C}(A_i \cdot A_j)},
\end{equation}
where $b_{ji}$ describes the influence from channel $i$ to channel $j$. Then, we use matrix multiplication between attention map $\mathbf{B}$ and the original feature map $\mathbf{A}$ to propagate channel information selectively. Again, we use another learnable scale parameter $\beta$ initialized from 0 to weight the element-wise summation between the aggregated feature map and the original feature map:
\begin{equation}
    D_j = \beta \sum_{i=1}^{C}(b_{ji}A_i) + A_j,
\end{equation}
In the above, the channel inter-dependencies are aggregated at each point.

\subsubsection{Point-wise Attention Module}
To retain the necessary local representation for localization, our classification network applies only two downsampling layer. In this case, the receptive field of this model is considerably restricted. The small receptive field makes it hard for the network to understand complex scenes. An excellent approach to enlarge the receptive field is PSPNet\cite{zhao2017pyramid}, which applies various scale of spatial pooling to gather global information. However, the variant sparsity of point cloud makes it hard to append different levels of pooling modules into the network. Inspired by PSANet\cite{zhao2018psanet}, we design a simplified point-wise spatial attention module to aggregate global information for each point. As shown in Figure \ref{fig:attention}(C), we calculate the spatial attention map using the same way as in Spatial Attention Module. Instead of summing the attention feature to the original feature, we directly concatenate the aggregated feature to the original feature map. Here, the aggregated global feature performs as a weighted global pooling module, which allows the resulting feature map to have both local and global feature. Thus, we can mine more region information from this module.

\subsection{Learning a Semantic Segmentation Network}
After acquiring the four distinct PCAMs, we use an element-wise maximum to get the max value at each position and upsample it to generate the pseudo mask. Then, to leverage the low-level contextual information and the pairwise relationship between points, we use the dCRF\cite{krahenbuhl2011efficient} to refine the pseudo labels.To provide a ready to use full-scale segmentation model, we retrain a segmentation network using the generated pseudo labels. Besides, deeper convolution neural networks have the ability to learn more feature representations despite some misclassified pseudo labels and produce better results. Here, we use the KPConv U-Net\cite{Ronneberger_2015} like structured segmentation model as our final model.

\begin{table*}[]
        \centering
        \setlength{\abovedisplayskip}{0pt}
        \setlength{\belowdisplayskip}{0pt}
        \footnotesize
        \setlength\tabcolsep{1pt}
        \begin{tabular}{l|cccccccccccccccccccc}
            \hline
              Supervision &wall &floor &cabinet & bed & chair &sofa & table &door &window &bookshelf & picture & counter &desk &curtain & fridge & showercurtain &toilet & sink & bathtub & other  \\
            \hline \hline
            Scene&97.3&  99.3& 46.9& 20.4&
            66.4& 23.2& 51.8& 72.7& 50.7&
            14.3& 26.7& 14.9& 27.2& 16.1&
            14.7& 8.8& 15.7& 26.8& 94.0& 74.0 \\
            Subcloud&77.6& 51.5& 17.8& 8.3&
       28.0& 8.6& 19.2& 24.7& 18.0& 5.6& 7.7& 5.1& 9.1& 5.8&
       3.7& 1.9& 2.6& 4.7& 1.7&20.9\\
           \hline

        \end{tabular}
        \caption{The class frequency (\%) on both scene-labels and subcloud-level labels.}
        \label{tab:freqlong}
        \vspace{-5pt}
    \end{table*}

\begin{table*}[]
        \centering
        \setlength{\abovedisplayskip}{0pt}
        \setlength{\belowdisplayskip}{0pt}
        \footnotesize
        \setlength\tabcolsep{1pt}
        \begin{tabular}{l|l|cccccccccccccccccccc|c}
            \hline
             Setting & Supervision &wall &floor &cabinet & bed & chair &sofa & table &door &window &B.S. & picture & cnt &desk &curtain & fridge & S.C. &toilet & sink & bathtub & other & mIoU \\
            \hline \hline
             PCAM(Baseline) & Scene & 54.9&48.3&14.1&34.7&32.9&45.3&26.1&0.6&3.3&46.5&0.6       &6.0&7.4&26.9&0.0&6.1&22.3&8.2&52.0&6.1&22.1\\
             MPRM(Ours) & Scene& 47.3&41.1&10.4&43.2&25.2&43.1&21.5&9.8&12.3&45.0&      9.0&13.9&21.1&40.9&1.8&29.4&14.3&9.2&39.9&10.0&24.4\\ 

             PCAM(Baseline) & Subcloud  &  59.0&53.8&24.7&64.9&45.7&60.7&42.8&31.5&37.0&55.9&31.0&12.0&      39.1&68.7&16.8&49.8&55.2&27.4&59.0&27.7&43.1\\ %
             MPRM(Ours) & Subcloud&56.1& 54.8& 32.0& 69.6& 49.5& 67.7& 46.6& 41.3& 44.2& 71.5& 28.3& 21.3& 49.2& 71.8& 38.1& 42.8& 43.6& 20.3& 49.0& 33.8 & 46.6\\
             \hline
             \multicolumn{21}{l}{dCRF post-processing:}\\
             \hline
             MPRM(Ours)&Subcloud &58.0& 57.3& 33.2& 71.8& 50.4& 69.8& 47.9& 42.1& 44.9& 73.8& 28.0& 21.5& 49.5& 72.0& 38.8& 44.1& 42.4& 20.0& 48.7& 34.4 & 47.4\\
            
            \hline

        \end{tabular}
        \caption{The class-specific segmentation results (mIoU) of pseudo labels on training set generated with different settings and different supervision levels. We only show the dCRF post-processed result for MPRM with subcloud-level supervision since we use this pseudo label to train our final segmentation model. (Here B.S. stands for bookshelf; S.C. stands for shower curtain; cnt stands for counter.)}
        \label{tab:detail}
        \vspace{-15pt}
    \end{table*}
\begin{table}[b]
\vspace{-10pt}
\centering
\begin{tabular}{@{}c|cccc|c|c@{}}
\hline
Fusion & PCAM & SA & CA & PSA & Training & Validation \\
\hline\hline
-      & $\surd$    &    &    &     & 44.3    & 39.3      \\
-      &      & $\surd$  &    &     & 44.8    & 39.4      \\
-      &      &    & $\surd$  &     & 44.3    & 39.3      \\
-      &      &    &    & $\surd$   & 44.7    & 39.5      \\
Max    & $\surd$    & $\surd$  &    &     & 46.0    & 40.3      \\
Max    & $\surd$    &    & $\surd$  &     & 45.9   & 40.0      \\
Max    & $\surd$    &    &    & $\surd$   & 45.6    & 40.4      \\
Max    & $\surd$    & $\surd$  & $\surd$  & $\surd$   & \textbf{46.6}    & \textbf{41.0}      \\
Sum    & $\surd$    & $\surd$  & $\surd$  & $\surd$   & 45.9    & 39.7       \\
\hline
\end{tabular}
\caption{The mIoU of pseudo labels with different paths and their combinations on training and validation set.}
\label{tab:mb_result}
\vspace{-10pt}
\end{table}

\begin{table*}[]
        \centering
        \setlength{\abovedisplayskip}{0pt}
        \setlength{\belowdisplayskip}{0pt}
        \footnotesize
        \setlength\tabcolsep{1pt}
        \begin{tabular}{l|c|cccccccccccccccccccc|c}
            \hline
             Setting & Retrain &wall &floor &cabinet & bed & chair &sofa & table &door &window &B.S. & picture & counter &desk &curtain & fridge & S.C. &toilet & sink & bathtub & other & mIoU \\
            \hline \hline
            
            \hline
             PCAM  & No&  57.3& 49.2& 20.4& 51.4& 44.3& 53.2& 37.1& 29.3& 32.4& 54.6& 20.5& 9.9& 34.8& 62.5& 10.0& 37.5& 55.3& 33.0& 49.5& 20.2& 38.1 \\
             MPRM  & No&  55.7& 50.7& 23.1& 57.5& 47.5& 53.5& 39.2& 32.6& 41.8& 63.6& 19.7& 19.2& 39.8& 66.3& 22.2& 44.1& 49.1& 23.4& 43.0& 28.5& 41.0 \\
             MPRM-CRF & No  &  55.0& 50.0& 23.5& 59.2& 47.6& 54.3& 41.7& 34.8& 41.2& 63.9& 20.6& 20.8& 40.6& 66.2& 24.1& 43.5& 48.5& 23.6& 41.9& 27.6& 41.4 \\
             MPRM-CRF& Yes&  59.4& 59.6& 25.1& 64.1& 55.7& 58.7& 45.6& 36.4& 40.3& 67.0& 16.1& 22.6& 42.9& 66.9& 24.1& 39.6& 47.0& 21.2& 44.7& 28.0& 43.2 \\
             \hline

        \end{tabular}
        \caption{The class-specific segmentation results (mIoU) on the validation set with subcloud level supervision. No retraining means directly using the output from the MPRM model as segmentation predictions. With retraining means that we retrain a full-scale segmentation network with our previous MPRM-CRF pseudo label.}
        \label{tab:retraining}
        \vspace{-10pt}
    \end{table*}

\begin{figure*}[ht]
\begin{center}
\includegraphics[width=1\linewidth]{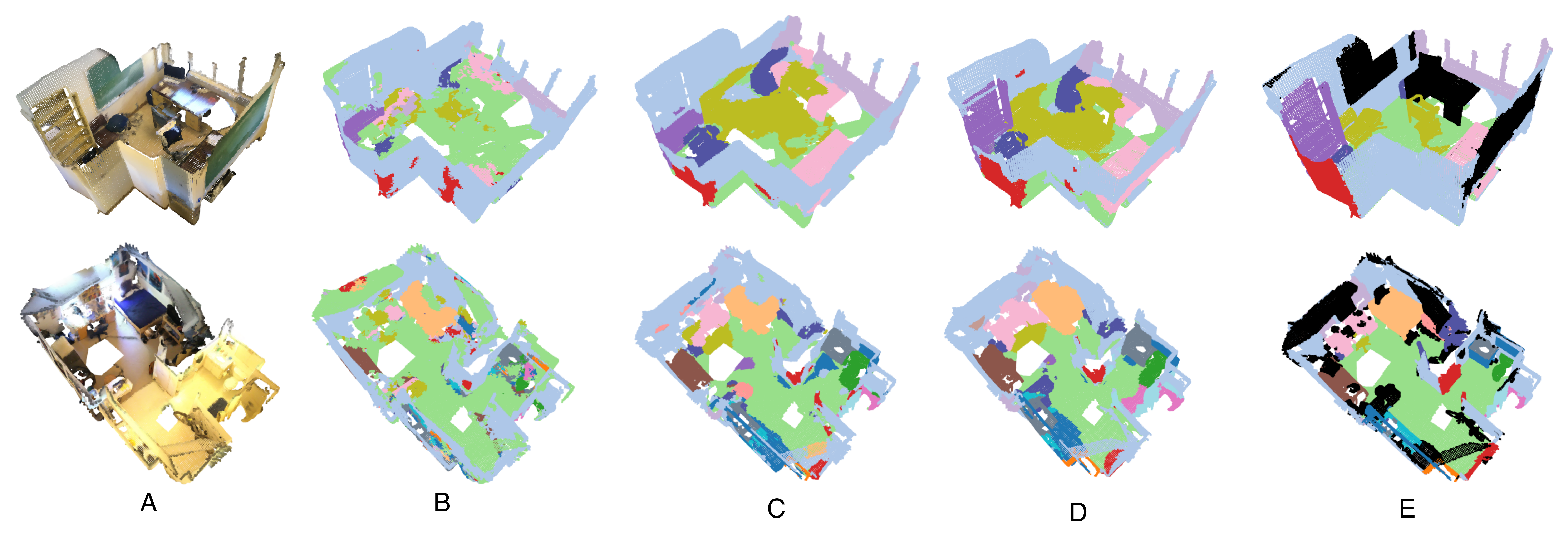}
\end{center}
   \caption{Visualizations of pseudo labels. (A)Input point clouds, (B)PCAMs trained with scene-level labels, (C)PCAMs trained with subcloud-level labels, (D)Multi-path region mining trained with subcloud-level labels, (E)Ground truth.}
\label{fig:stage}
\vspace{-10pt}
\end{figure*}

\section{Implementation Details}
\subsection{Dataset and Weak Labels}
We perform our experiments on ScanNet\cite{dai2017scannet} dataset. ScanNet has 1513 training scenes and 100 test scenes with 20 classes. We adopt the official train-val split, where there are 1205 training scenes and 312 validation scenes. We first preprocess the ScanNet data following the approach of \cite{thomas2019kpconv}. Then, for scene-level labels, we label each scene with the classes that appeared in each scene; for subcloud-level labels, we use a query radius $\mathnormal{r}=2.0m$ to sample subclouds. The average size of the training scenes is $5.5\times5.1\times2.4 m$. The resulting average subcloud number per scene for the training set is 18.4, and we label each subcloud with the classes that appeared in the subcloud. 

\subsection{Generating Pseudo-labels}\label{sec:masksetting}
Our classification backbone is a simpler version of KPConv\cite{thomas2019kpconv} classification network. We use a simple convolution layer followed by 5 ResNet\cite{he2016deep} bottleneck blocks, among which, we use deformable kernels in the last three blocks. The second and fourth ResNet blocks apply strided convolution as the downsampling layers.
We take a four-dimensional input feature based on the RGB value and an indicator $1$ in case the point is black. Due to the network capacity, we cannot feed a whole scene input the network while training with scene-level labels. For fair comparison, we randomly sample subclouds using the same radius and append scene-level labels to them. The input data is formed with subclouds stacked along the batch dimension, as the convolution and pooling operation is solely based on a neighboring index generated using KD-Tree. As the number of points varies a lot in each subcloud, there is no fixed batch size. A batch is formed by stacking subclouds until the batch limit number of points are met. The batch limit is chosen to accord with the target batch size. We use a dropout layer before each $1\times1$ convolution classifier with a dropout rate of $0.5$.
During training, we use a Momentum SGD optimizer, with a momentum of 0.98 and an initial learning rate of 0.01. We apply the exponential learning rate decay to ensure it is divided by 10 every 100 epochs. The target batch size is set at 10, and the network converges at about 400 epochs.

\subsection{Training Segmentation Model}
We use the KPConv\cite{thomas2019kpconv} segmentation model as our final segmentation model. The model consists of four blocks, and each block is composed of two deformable ResNet bottleneck block and a strided convolution layer for downsampling. We use the same training setting as \ref{sec:masksetting}. The model converges at around 200 epochs.

\section{Experiments}

\subsection{Scene-level Versus Subcloud-level Labels}\label{subexp}
In this section, we analyze our scene- and subcloud-level labels. As we discussed in \ref{sec:weaklabel}, there are two main challenges for training with scene-level label. In Table \ref{tab:freqlong}, we show the frequency of each class in both scene- and subcloud-level labels. We can observe that classes like floors, walls are almost inevitable in every scene. Through our subsampling, half of the subclouds do not contain the class floor since we are sampling at different heights. Although walls are hard to avoid, we still produced more than 20\% data as negative samples, which provides negative samples for the dominant classes. The effect is more obvious when it comes to frequent small objects like table, chair, and doors. Thus, the network can learn better discriminative features about those classes. Although some classes like shower curtain and bathtub seem to be rare in subcloud-level samples in percentage, the actual number of samples is increased.

To illustrate the effect, we compare both our PCAM baseline and the MPRM with scene- and subcloud-level labels. Table \ref{tab:detail} shows class-specific pseudo label performance. The performances of using subcloud-level labels are leading by a large margin on both settings. Specifically, we can observe that using scene-level labels, the segmentation performances on small objects are severely inferior, especially those objects usually placed near walls. This phenomenon accords with our previous assumption that the network cannot learn discriminative features for dominant classes and simply generate high scores for those classes on every point. We can clearly observe this from Figure \ref{fig:stage}(B), that the network tends to predict every point as walls and floors. Therefore, we prove that using subcloud-level label can significantly relieve the class imbalance problem in scene-level labels.

\subsection{Pseudo Label Evaluation}
As shown in Table \ref{tab:detail}, we present the class-specific segmentation result of our pseudo label on the training set. Results show that the Multi-path Region Mining module can help increase the segmentation performance using both scene-level and subcloud-level labels. With scene-level supervision, we observe our baseline method can hardly find anything about the classes that are near the dominant classes like door, window, picture, with our MPRM module the performance on these classes increased by a large margin. It shows that with our MPRM, the network learns to separate small objects from dominant classes. From Figure \ref{fig:stage}, it can be seen that MPRM generates more small object region from dominant classes and produce better smoothness over the space.

We also present the segmentation result of MPRM post-processed by a dCRF\cite{krahenbuhl2011efficient}, which incorporate low-level features like color and pairwise smoothness information to refine the pseudo label. Then, we use the MP-CRF result as the pseudo per point label to train our full-scale segmentation network.

\subsection{Ablation Study}
In this section, we conduct detailed experiments to evaluate our Multi-path Region Mining Module. As shown in Table \ref{tab:mb_result}, we evaluate the performance of each path and their combination with the original PCAM. Also, we compare two different fusion methods. Note that the network is trained simultaneously using the losses from all four paths. Comparing with the baseline results, our PCAM path within MPRM performs better than merely training one branch. Thus, we can learn that the losses from different paths indeed help the classification backbone to produce better features. Among the four paths, the spatial attention path performs best alone, and all three modules produce better results than the original PCAM. Besides, the combination of each two branches outperforms their branches solely, which proves that our different paths are indeed learning various features from the classification network. As the four paths merging result is higher than any other combination, we show that all the four paths help produce better pseudo labels. 

We compare the element-wise maximum fusion with the element-wise sum fusion in the last two rows. It is evident that max fusion is better. As we use global average pooling to get our classification predictions, the score amplitude in each class may vary a lot. Using max fusion, we can keep the fusion within each class. However, using sum fusion may cause some overwhelming classes, thus influence the final classification result.

\begin{table}[]
\begin{center}
\setlength\tabcolsep{20pt}
\begin{tabular}{l|c}
\hline
Method & mIoU \\
\hline\hline
Supervision: Point-level\\
Pointnet++\cite{qi2017pointnet++}  & 33.9 \\
SPLATNet\cite{su2018splatnet} & 39.3 \\
TangentConv\cite{tatarchenko2018tangent} & 43.8\\
PointCNN\cite{li2018pointcnn} &  45.8\\
KPConv\cite{thomas2019kpconv} &  68.4\\
SubSparseCNN\cite{graham20183d} &  72.5\\
\hline
Supervision: Subcloud-level\\
Ours  & 41.1\\
\hline
\end{tabular}
\end{center}
\caption{3D scene semantic segmentation results (mIoU) on ScanNet test set. The results are taken from the online benchmark.}
\label{tabel:result}
\vspace{-15pt}
\end{table}

\begin{figure}[h]
\begin{center}
   \includegraphics[width=0.8\linewidth]{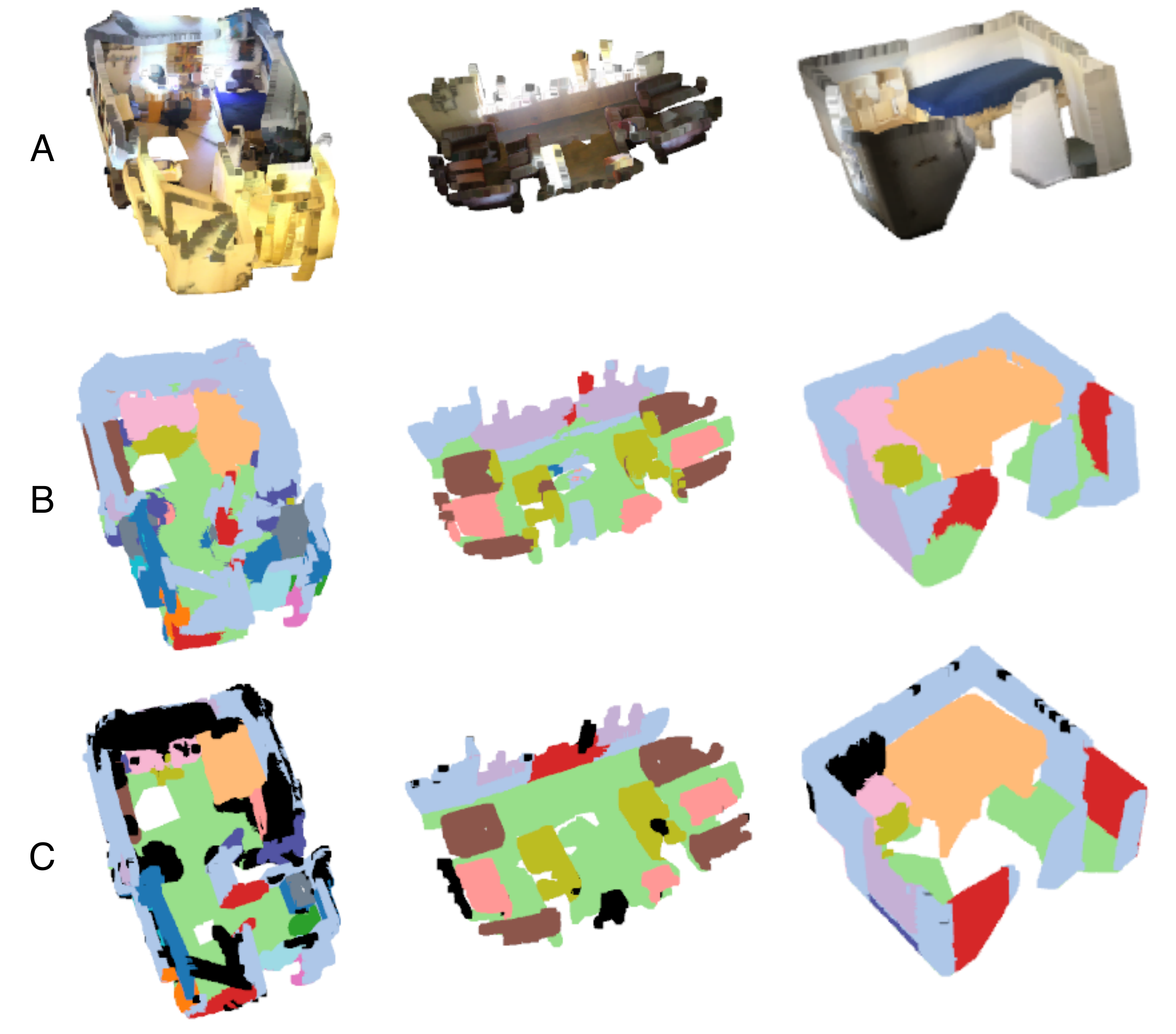}
\end{center}
   \caption{Qualitative results of our final segmentation model. (A) Input point clouds, (B) Segmentation predictions, (C) Ground truth. Note that the black color in the ground truth indicates unclassified points, which are ignored during evaluation.}
\label{fig:vis_result}
\vspace{-20pt}
\end{figure}

\subsection{Segmentation Results}
In Table \ref{tab:retraining}, we show the segmentation performance with subcloud-level annotation on the validation set. We can observe that MPRM outperforms the baseline method on the validation set. We also see that retraining a full-scale segmentation network can provide better results than the raw feature maps as a deeper network can learn more representation than our shallow network. Besides, by retraining, we can aggregate the low level feature generated from dCRF to our final model, and incorporate the post-processing step into an end-to-end network.
We compare our final result with some fully supervised approaches in Table \ref{tabel:result}. There is still a huge performance gap between our weakly supervised approach and the current state-of-the-art methods, but our weakly supervised approach is compatible with some fully supervised methods. We also show qualitative results in Figure \ref{fig:vis_result}.

\vspace{-10pt}
\section{Conclusions}
\vspace{-5pt}
In this paper, we have proposed a weakly supervised learning scheme for 3D point cloud scene semantic segmentation. Our subcloud labeling strategy can significantly reduce the labor and time cost for annotation on 3D datasets. Besides, we propose a Multi-path Region Mining Module to improve segmentation performance. The result of our weakly supervised approach is compatible with some fully supervised methods.\\
\textbf{Acknowledgement} This work was conducted within the Delta-NTU Corporate Lab for Cyber-Physical Systems with funding support from Delta Electronics Inc. and the National Research Foundation (NRF) Singapore under the Corp Lab@University Scheme. This work is also partly supported by the National Research Foundation Singapore under its AI Singapore Programme (Award Number: AISG-RP-2018-003) and the MOE Tier-1 research grant: RG22/19 (S).

{\small
\bibliographystyle{ieee_fullname}
\bibliography{egpaper}

\begin{thebibliography}{10}\itemsep=-1pt

\bibitem{ahn2018learning}
Jiwoon Ahn and Suha Kwak.
\newblock Learning pixel-level semantic affinity with image-level supervision
  for weakly supervised semantic segmentation.
\newblock In {\em Proceedings of the IEEE Conference on Computer Vision and
  Pattern Recognition}, pages 4981--4990, 2018.

\bibitem{atzmon2018point}
Matan Atzmon, Haggai Maron, and Yaron Lipman.
\newblock Point convolutional neural networks by extension operators.
\newblock {\em arXiv preprint arXiv:1803.10091}, 2018.

\bibitem{bearman2016s}
Amy Bearman, Olga Russakovsky, Vittorio Ferrari, and Li Fei-Fei.
\newblock What’s the point: Semantic segmentation with point supervision.
\newblock In {\em European conference on computer vision}, pages 549--565.
  Springer, 2016.

\bibitem{ben20183dmfv}
Yizhak Ben-Shabat, Michael Lindenbaum, and Anath Fischer.
\newblock 3dmfv: Three-dimensional point cloud classification in real-time
  using convolutional neural networks.
\newblock {\em IEEE Robotics and Automation Letters}, 3(4):3145--3152, 2018.

\bibitem{boulch2017unstructured}
Alexandre Boulch, Bertrand Le~Saux, and Nicolas Audebert.
\newblock Unstructured point cloud semantic labeling using deep segmentation
  networks.
\newblock {\em 3DOR}, 2:7, 2017.

\bibitem{cai2019deep}
Jun-Xiong Cai, Tai-Jiang Mu, Yu-Kun Lai, and Shi-Min Hu.
\newblock Deep point-based scene labeling with depth mapping and geometric
  patch feature encoding.
\newblock {\em Graphical Models}, 104:101033, 2019.

\bibitem{choy20194d}
Christopher Choy, JunYoung Gwak, and Silvio Savarese.
\newblock 4d spatio-temporal convnets: Minkowski convolutional neural networks.
\newblock {\em arXiv preprint arXiv:1904.08755}, 2019.

\bibitem{dai2017scannet}
Angela Dai, Angel~X. Chang, Manolis Savva, Maciej Halber, Thomas Funkhouser,
  and Matthias Nie{\ss}ner.
\newblock Scannet: Richly-annotated 3d reconstructions of indoor scenes.
\newblock In {\em Proc. Computer Vision and Pattern Recognition (CVPR), IEEE},
  2017.

\bibitem{fan2018associating}
Ruochen Fan, Qibin Hou, Ming-Ming Cheng, Gang Yu, Ralph~R Martin, and Shi-Min
  Hu.
\newblock Associating inter-image salient instances for weakly supervised
  semantic segmentation.
\newblock In {\em Proceedings of the European Conference on Computer Vision
  (ECCV)}, pages 367--383, 2018.

\bibitem{fu2019dual}
Jun Fu, Jing Liu, Haijie Tian, Yong Li, Yongjun Bao, Zhiwei Fang, and Hanqing
  Lu.
\newblock Dual attention network for scene segmentation.
\newblock In {\em Proceedings of the IEEE Conference on Computer Vision and
  Pattern Recognition}, pages 3146--3154, 2019.

\bibitem{graham20183d}
Benjamin Graham, Martin Engelcke, and Laurens van~der Maaten.
\newblock 3d semantic segmentation with submanifold sparse convolutional
  networks.
\newblock In {\em Proceedings of the IEEE Conference on Computer Vision and
  Pattern Recognition}, pages 9224--9232, 2018.

\bibitem{groh2018flex}
Fabian Groh, Patrick Wieschollek, and Hendrik~PA Lensch.
\newblock Flex-convolution.
\newblock In {\em Asian Conference on Computer Vision}, pages 105--122.
  Springer, 2018.

\bibitem{he2016deep}
Kaiming He, Xiangyu Zhang, Shaoqing Ren, and Jian Sun.
\newblock Deep residual learning for image recognition.
\newblock In {\em Proceedings of the IEEE conference on computer vision and
  pattern recognition}, pages 770--778, 2016.

\bibitem{hou20193d}
Ji Hou, Angela Dai, and Matthias Nie{\ss}ner.
\newblock 3d-sis: 3d semantic instance segmentation of rgb-d scans.
\newblock In {\em Proceedings of the IEEE Conference on Computer Vision and
  Pattern Recognition}, pages 4421--4430, 2019.

\bibitem{hu2018semantic}
Shi-Min Hu, Jun-Xiong Cai, and Yu-Kun Lai.
\newblock Semantic labeling and instance segmentation of 3d point clouds using
  patch context analysis and multiscale processing.
\newblock {\em IEEE transactions on visualization and computer graphics}, 2018.

\bibitem{hua2018pointwise}
Binh-Son Hua, Minh-Khoi Tran, and Sai-Kit Yeung.
\newblock Pointwise convolutional neural networks.
\newblock In {\em Proceedings of the IEEE Conference on Computer Vision and
  Pattern Recognition}, pages 984--993, 2018.

\bibitem{huang2018weakly}
Zilong Huang, Xinggang Wang, Jiasi Wang, Wenyu Liu, and Jingdong Wang.
\newblock Weakly-supervised semantic segmentation network with deep seeded
  region growing.
\newblock In {\em Proceedings of the IEEE Conference on Computer Vision and
  Pattern Recognition}, pages 7014--7023, 2018.

\bibitem{khoreva2017simple}
Anna Khoreva, Rodrigo Benenson, Jan Hosang, Matthias Hein, and Bernt Schiele.
\newblock Simple does it: Weakly supervised instance and semantic segmentation.
\newblock In {\em Proceedings of the IEEE conference on computer vision and
  pattern recognition}, pages 876--885, 2017.

\bibitem{kolesnikov2016seed}
Alexander Kolesnikov and Christoph~H Lampert.
\newblock Seed, expand and constrain: Three principles for weakly-supervised
  image segmentation.
\newblock In {\em European Conference on Computer Vision}, pages 695--711.
  Springer, 2016.

\bibitem{krahenbuhl2011efficient}
Philipp Kr{\"a}henb{\"u}hl and Vladlen Koltun.
\newblock Efficient inference in fully connected crfs with gaussian edge
  potentials.
\newblock In {\em Advances in neural information processing systems}, pages
  109--117, 2011.

\bibitem{li2018so}
Jiaxin Li, Ben~M Chen, and Gim Hee~Lee.
\newblock So-net: Self-organizing network for point cloud analysis.
\newblock In {\em Proceedings of the IEEE conference on computer vision and
  pattern recognition}, pages 9397--9406, 2018.

\bibitem{li2018pointcnn}
Yangyan Li, Rui Bu, Mingchao Sun, Wei Wu, Xinhan Di, and Baoquan Chen.
\newblock Pointcnn: Convolution on x-transformed points.
\newblock In {\em Advances in Neural Information Processing Systems}, pages
  820--830, 2018.

\bibitem{lin2016scribblesup}
Di Lin, Jifeng Dai, Jiaya Jia, Kaiming He, and Jian Sun.
\newblock Scribblesup: Scribble-supervised convolutional networks for semantic
  segmentation.
\newblock In {\em Proceedings of the IEEE Conference on Computer Vision and
  Pattern Recognition}, pages 3159--3167, 2016.

\bibitem{liu2019point2sequence}
Xinhai Liu, Zhizhong Han, Yu-Shen Liu, and Matthias Zwicker.
\newblock Point2sequence: Learning the shape representation of 3d point clouds
  with an attention-based sequence to sequence network.
\newblock In {\em Proceedings of the AAAI Conference on Artificial
  Intelligence}, volume~33, pages 8778--8785, 2019.

\bibitem{mao2019interpolated}
Jiageng Mao, Xiaogang Wang, and Hongsheng Li.
\newblock Interpolated convolutional networks for 3d point cloud understanding.
\newblock In {\em Proceedings of the IEEE International Conference on Computer
  Vision}, pages 1578--1587, 2019.

\bibitem{maturana2015voxnet}
Daniel Maturana and Sebastian Scherer.
\newblock Voxnet: A 3d convolutional neural network for real-time object
  recognition.
\newblock In {\em 2015 IEEE/RSJ International Conference on Intelligent Robots
  and Systems (IROS)}, pages 922--928. IEEE, 2015.

\bibitem{qi2017pointnet}
Charles~R Qi, Hao Su, Kaichun Mo, and Leonidas~J Guibas.
\newblock Pointnet: Deep learning on point sets for 3d classification and
  segmentation.
\newblock In {\em Proceedings of the IEEE Conference on Computer Vision and
  Pattern Recognition}, pages 652--660, 2017.

\bibitem{qi2017pointnet++}
Charles~Ruizhongtai Qi, Li Yi, Hao Su, and Leonidas~J Guibas.
\newblock Pointnet++: Deep hierarchical feature learning on point sets in a
  metric space.
\newblock In {\em Advances in neural information processing systems}, pages
  5099--5108, 2017.

\bibitem{Ronneberger_2015}
Olaf Ronneberger, Philipp Fischer, and Thomas Brox.
\newblock U-net: Convolutional networks for biomedical image segmentation.
\newblock {\em Medical Image Computing and Computer-Assisted Intervention –
  MICCAI 2015}, page 234–241, 2015.

\bibitem{sauder2019selfsupervised}
Jonathan Sauder and Bjarne Sievers.
\newblock Self-supervised deep learning on point clouds by reconstructing
  space, 2019.

\bibitem{sensor2016occipital}
Structure Sensor.
\newblock Occipital inc, 2016.

\bibitem{song2019box}
Chunfeng Song, Yan Huang, Wanli Ouyang, and Liang Wang.
\newblock Box-driven class-wise region masking and filling rate guided loss for
  weakly supervised semantic segmentation.
\newblock In {\em Proceedings of the IEEE Conference on Computer Vision and
  Pattern Recognition}, pages 3136--3145, 2019.

\bibitem{su2018splatnet}
Hang Su, Varun Jampani, Deqing Sun, Subhransu Maji, Evangelos Kalogerakis,
  Ming-Hsuan Yang, and Jan Kautz.
\newblock Splatnet: Sparse lattice networks for point cloud processing.
\newblock In {\em Proceedings of the IEEE Conference on Computer Vision and
  Pattern Recognition}, pages 2530--2539, 2018.

\bibitem{su2015multi}
Hang Su, Subhransu Maji, Evangelos Kalogerakis, and Erik Learned-Miller.
\newblock Multi-view convolutional neural networks for 3d shape recognition.
\newblock In {\em Proceedings of the IEEE international conference on computer
  vision}, pages 945--953, 2015.

\bibitem{tatarchenko2018tangent}
Maxim Tatarchenko, Jaesik Park, Vladlen Koltun, and Qian-Yi Zhou.
\newblock Tangent convolutions for dense prediction in 3d.
\newblock In {\em Proceedings of the IEEE Conference on Computer Vision and
  Pattern Recognition}, pages 3887--3896, 2018.

\bibitem{thabet2019mortonnet}
Ali Thabet, Humam Alwassel, and Bernard Ghanem.
\newblock Mortonnet: Self-supervised learning of local features in 3d point
  clouds, 2019.

\bibitem{thomas2019kpconv}
Hugues Thomas, Charles~R Qi, Jean-Emmanuel Deschaud, Beatriz Marcotegui,
  Fran{\c{c}}ois Goulette, and Leonidas~J Guibas.
\newblock Kpconv: Flexible and deformable convolution for point clouds.
\newblock {\em arXiv preprint arXiv:1904.08889}, 2019.

\bibitem{wang2019weakly3d}
Haiyan Wang, Xuejian Rong, Liang Yang, Shuihua Wang, and Yingli Tian.
\newblock Towards weakly supervised semantic segmentation in 3d
  graph-structured point clouds of wild scenes.
\newblock In {\em Proceedings of the British Machine Vision Conference}, 2019.

\bibitem{wang2018non}
Xiaolong Wang, Ross Girshick, Abhinav Gupta, and Kaiming He.
\newblock Non-local neural networks.
\newblock In {\em Proceedings of the IEEE Conference on Computer Vision and
  Pattern Recognition}, pages 7794--7803, 2018.

\bibitem{wei2017object}
Yunchao Wei, Jiashi Feng, Xiaodan Liang, Ming-Ming Cheng, Yao Zhao, and
  Shuicheng Yan.
\newblock Object region mining with adversarial erasing: A simple
  classification to semantic segmentation approach.
\newblock In {\em Proceedings of the IEEE conference on computer vision and
  pattern recognition}, pages 1568--1576, 2017.

\bibitem{xu2018spidercnn}
Yifan Xu, Tianqi Fan, Mingye Xu, Long Zeng, and Yu Qiao.
\newblock Spidercnn: Deep learning on point sets with parameterized
  convolutional filters.
\newblock In {\em Proceedings of the European Conference on Computer Vision
  (ECCV)}, pages 87--102, 2018.

\bibitem{zhang2018self}
Han Zhang, Ian Goodfellow, Dimitris Metaxas, and Augustus Odena.
\newblock Self-attention generative adversarial networks.
\newblock {\em arXiv preprint arXiv:1805.08318}, 2018.

\bibitem{zhao2017pyramid}
Hengshuang Zhao, Jianping Shi, Xiaojuan Qi, Xiaogang Wang, and Jiaya Jia.
\newblock Pyramid scene parsing network.
\newblock In {\em Proceedings of the IEEE conference on computer vision and
  pattern recognition}, pages 2881--2890, 2017.

\bibitem{zhao2018psanet}
Hengshuang Zhao, Yi Zhang, Shu Liu, Jianping Shi, Chen Change~Loy, Dahua Lin,
  and Jiaya Jia.
\newblock Psanet: Point-wise spatial attention network for scene parsing.
\newblock In {\em Proceedings of the European Conference on Computer Vision
  (ECCV)}, pages 267--283, 2018.

\bibitem{zhou2016learning}
Bolei Zhou, Aditya Khosla, Agata Lapedriza, Aude Oliva, and Antonio Torralba.
\newblock Learning deep features for discriminative localization.
\newblock In {\em Proceedings of the IEEE conference on computer vision and
  pattern recognition}, pages 2921--2929, 2016.

\end{thebibliography}
}

\end{document}